%% file: main.tex
\documentclass[runningheads]{llncs}
\usepackage[T1]{fontenc}
\usepackage[
backend=biber,
style=ieee,
doi=false,isbn=false,url=false,eprint=false
]{biblatex}
\usepackage[precision=2, unit=mm]{lengthconvert}
\usepackage{tikz}
\usetikzlibrary{patterns}

\usepackage{pdfpages}
\usepackage{graphicx}
\usepackage{booktabs}
\usepackage{amsmath}
\usepackage{amssymb}
\usepackage{array}
\usepackage{xcolor}
\usepackage{colortbl}
\usepackage{makecell}
\usepackage{multirow}
\usepackage{float}
\usepackage{hyperref}
\usepackage{pifont}
\graphicspath{ {./figures/} }
\begin{document}
\def\methodName{HieraSurg}
\def\S2M{S2M} 
\def\M2V{M2V} 
\title{\methodName{}: Hierarchy-Aware Diffusion Model for Surgical Video Generation}
%
%
\author{Diego Biagini\inst{1,2} \and Nassir Navab \inst{1,2} \and Azade Farshad \inst{1,2}}

%
%
\institute{\textsuperscript{1}Chair for Computer Aided Medical Procedures (CAMP), TU Munich, Germany \\ \textsuperscript{2}Munich Center for Machine Learning (MCML) \\ \email{\{diego.biagini,azade.farshad\}@tum.de}}

%
%
%
\maketitle              
\input{chapters/0_abstract}
\input{chapters/1_intro}

\input{chapters/2_method}
\input{chapters/3_results}

\input{chapters/4_conclusion}


%
%
%
%
\clearpage
\printbibliography
\end{document}

%% file: chapters/0_abstract.tex
\begin{abstract}
Surgical Video Synthesis has emerged as a promising research direction following the success of diffusion models in general-domain video generation. Although existing approaches achieve high-quality video generation, most are unconditional and fail to maintain consistency with surgical actions and phases, lacking the surgical understanding and fine-grained guidance necessary for factual simulation. We address these challenges by proposing \textbf{\methodName{}}, a hierarchy-aware surgical video generation framework consisting of two specialized diffusion models. Given a surgical phase and an initial frame, \methodName{} first predicts future coarse-grained semantic changes through a segmentation prediction model. The final video is then generated by a second-stage model that augments these temporal segmentation maps with fine-grained visual features, leading to effective texture rendering and integration of semantic information in the video space. Our approach leverages surgical information at multiple levels of abstraction, including surgical phase, action triplets, and panoptic segmentation maps. The experimental results on Cholecystectomy Surgical Video Generation demonstrate that the model significantly outperforms prior work both quantitatively and qualitatively, showing strong generalization capabilities and the ability to generate higher frame-rate videos. The model exhibits particularly fine-grained adherence when provided with existing segmentation maps, suggesting its potential for practical surgical applications. \\ \textbf{Project Webpage}: \href{https://diegobiagini.github.io/HieraSurg/}{diegobiagini.github.io/HieraSurg/}


\end{abstract}

%% file: chapters/1_intro.tex
\begin{figure}[h]
\centering
\includegraphics[width=0.92\textwidth]{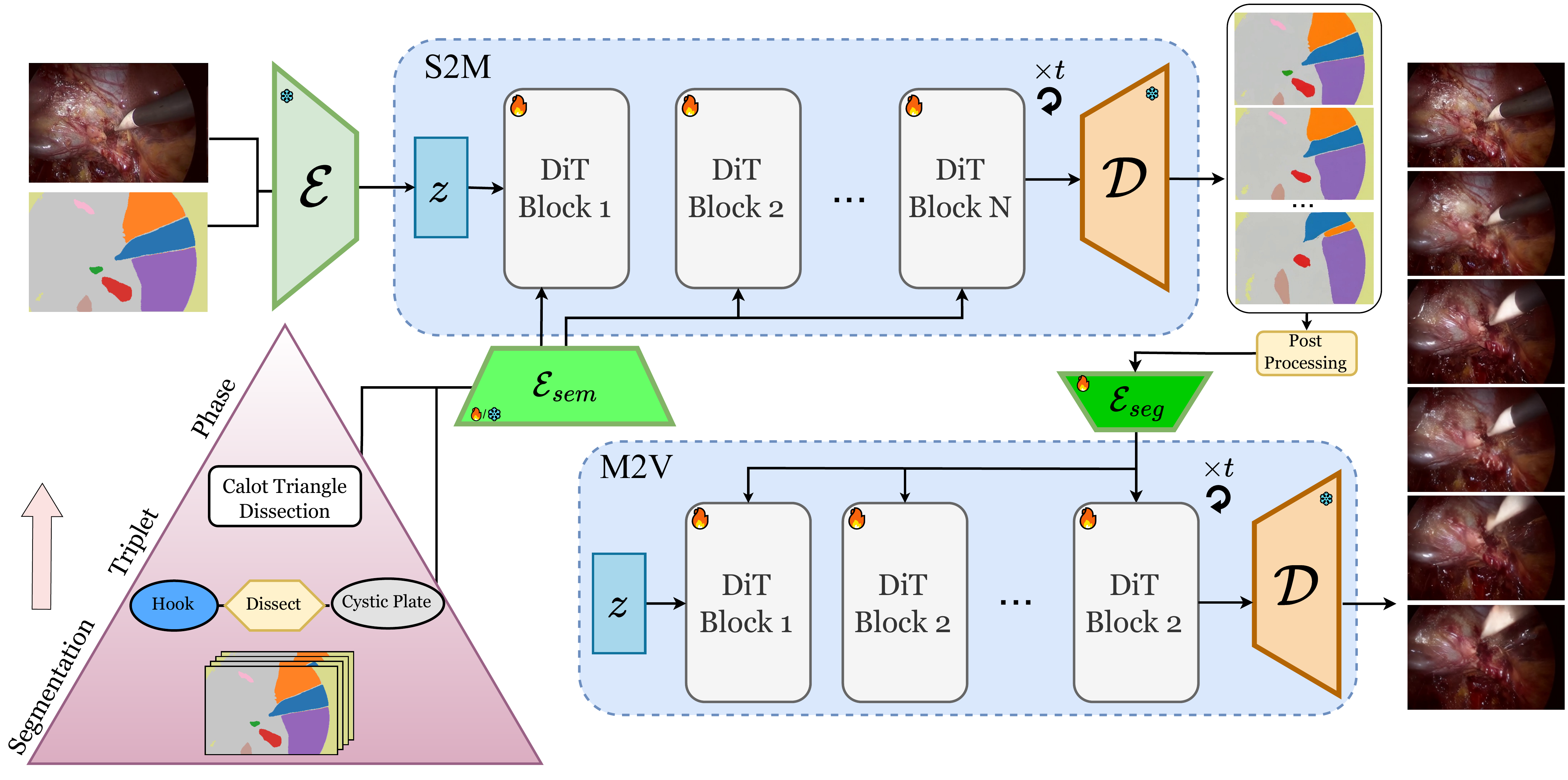}
\caption{\textbf{HieraSurg Pipeline}. \textbf{Left}: we take the hierarchical representation of a surgical scene, \textbf{Right}: the inference pipeline and components of HieraSurg.  \S2M{} takes an initial segmentation map and predicts the future evolution of the surgical scene by considering phase and triplet information. Finally, the output of \S2M{} is fed into \M2V{} to generate a video according to the predicted segmentation.}
\label{fig:hierasurg}
\end{figure}

\section{Introduction}
Recent advances in model efficiency, diffusion modelling and data curation have paved the way for generating videos closely mimicking reality with high fidelity \cite{blattmann2023stable,yang2025cogvideox,kong_hunyuanvideo_2025}. These developments indicate that such models are not merely replicating data-driven correlations; rather, they develop an internal world model, exhibiting emergent capabilities that approximate those of a simulator \cite{kang2024far}. 

In the medical domain, a natural question arises: can video generative models reliably function within the constrained environment of surgery, where implicit rules govern scene evolution? In this work, we evaluate the potential of these models as predictive tools for anticipating the short-term evolution of surgical scenes. Moreover, this study underscores another vital application of generative models in Surgical Data Science, namely, mitigating data scarcity in low-data regimes \cite{saragih2024using}. Previous works exploring generative video models in the surgical domain are Endora \cite{li_endora_2024}, an unconditional video model that incorporates semantic features from the DINO \cite{caron2021emerging} backbone into the training process, and VISAGE \cite{yeganeh_visage_2025} in which video generation is conditioned on action triplets after being embedded through the CLIP \cite{radford2021learning} encoder.
However, most of these works only consider one of the facets of the surgical scene. In particular, we posit that it's possible to describe a surgical scene in a hierarchical way (\textit{cf.} 
 \autoref{fig:hierasurg}), where information can be structured at different abstraction levels. We propose that a generative process should thus be split along the vertical axis of the pyramid, with certain components being tailored for the generation of a particular layer of the hierarchy. This is the central idea behind \textbf{HieraSurg} and its components: a first-stage model, \textbf{HieraSurg-\S2M{} (Semantic to Map)}, takes in input the high-level surgical information of the scene to generate a set of panoptic segmentation maps, which are used to guide the generation of a video by the second-stage model, \textbf{HieraSurg-\M2V{}  (Map to Video)}.
In training the second stage model, a large number of segmentation maps are required to learn the relationship between semantic space and video space. Due to data scarcity, we devise an automated labelling pipeline based on Segment Anything 2 (SAM2) \cite{ravi_sam_2024} to extract panoptic segmentation maps from unlabeled surgical videos.

We validate our framework on laparoscopic surgeries from Cholec80 \cite{twinanda_endonet_2017} and  CholecT45 \cite{nwoye_rendezvous_2022} datasets. Our main contributions are as follows: 1) \textbf{\methodName{}}, a coupled pair of generative video models that can synthesize realistic laparoscopic videos with unprecedented visual quality and frame rate; 2) an automatic panoptic segmentation pipeline tailored to low frame-rate surgical videos; and 3) extensive experiments assessing the visual quality, motion fidelity, and short-term scene prediction performance of HieraSurg using both fidelity and reconstruction / detection-based metrics.

%% file: chapters/2_method.tex
\vspace{-5pt}
\section{Method}
\vspace{5pt} \noindent \textbf{Definitions}
In a given dataset of videos, each video is a sequence $V = {x_i,..., x_F}$ comprised of $F$ frames of resolution $H$ by $W$. We denote the encoder mapping a video to a lower-dimensional latent space as $\mathcal{E}$, a Variational Autoencoder (VAE) following Latent Diffusion Models (LDMs) \cite{rombach_high-resolution_2022}. 
The mapping from latent space back to image space is carried out by the decoder $\mathcal{D}$.
A panoptic segmentation map for a video is a tensor $x\in \mathbb{N}^{F\times H\times W}$, where an integer value or a background value is assigned to each pixel, denoting the entity found there. 
From here on, we will refer to panoptic segmentation maps as segmentation maps for brevity.

\vspace{5pt} \noindent \textbf{Diffusion Models}\label{sec:vdm}
Diffusion models allow data generation by gradually denoising Gaussian samples to produce elements that match a given distribution through the forward and backward diffusion process. The forward process adds Gaussian noise to a data sample $\mathbf{x}_0$ through $T$ timesteps, producing a sequence of increasingly noisy samples ${\mathbf{x}_1, ..., \mathbf{x}_T}$. At each timestep $t$, noise is added according to
$q(\mathbf{x}_t|\mathbf{x}_{t-1}) = \mathcal{N}(\mathbf{x}_t; \sqrt{1-\beta_t}\mathbf{x}_{t-1}, \beta_t\mathbf{I})$,
where ${\beta_t}_{t=1}^T$ is a fixed noise schedule and $\mathcal{N}(\mu, \sigma^2)$ denotes a Gaussian distribution with mean $\mu$ and variance $\sigma^2$. The reverse process aims to learn how to denoise a sample by computing the posterior $q(\mathbf{x}_{t-1}|\mathbf{x}_t)$, which is intractable and requires approximate methods. In the standard formulation introduced in \cite{ho_denoising_2020}, a neural network $\epsilon_\theta$ predicts the noise component at each step. The reverse process can be expressed as:
$$ p_\theta(\mathbf{x}_{t-1}|\mathbf{x}_t) = \mathcal{N}(\mathbf{x}_{t-1}; \mu_\theta(\mathbf{x}_t, t), \sigma_t^2\mathbf{I}) $$
where $\mu_\theta(\mathbf{x}_t, t)$ is derived from $\epsilon_\theta(x_t,t)$.
The denoiser model is trained by optimizing a variational lower bound on the log-likelihood, simplifying the objective:
$$\mathcal{L} = \mathbb{E}_{t,\mathbf{x}_0,\epsilon}\left[|\epsilon - \epsilon_\theta(\mathbf{x}_t, t)|_2^2\right]$$
where $\epsilon$ is the random noise added in the forward process and $\mathbf{x}_t$ is obtained by:
$\mathbf{x}_t = \sqrt{\alpha_t}\mathbf{x}_0 + \sqrt{1-\alpha_t}\epsilon$
with $\alpha_t = \prod_{i=1}^t(1-\beta_i)$.
During sampling, new data is generated by starting with random Gaussian noise $\mathbf{x}_T \sim \mathcal{N}(0, \mathbf{I})$ and iteratively applying the learned reverse process for $T$ steps by predicting and removing noise components until a clean sample $\mathbf{x}_0$ is obtained.


\vspace{5pt} \noindent \textbf{Video Segmentation}\label{sec:videosegmap}
In our video generation pipeline, we rely heavily on segmentation maps as an intermediate representation. Obtaining such labels for surgical data is costly due to the length of recordings, and existing datasets like CholecSeg8k \cite{hong_cholecseg8k_2020} and Endoscapes \cite{murali2023endoscapes} are too limited. Given the data-intensive nature of diffusion models, an automatic labeling pipeline is needed.
Although some works have addressed this problem \cite{li2024critical}, none fully meet our specific needs. We therefore developed a pipeline that leverages generalist segmentation models-namely, SAM and SAM2 \cite{kirillov2023segment,ravi_sam_2024}—and robust feature extractors. Initially, we used DINO \cite{caron2021emerging} as our feature extractor; however, RADIO \cite{ranzinger_am-radio_2024} produced superior results due to its native support for non-square images and feature representations more compatible with SAM.
For each frame, we extract segmentation maps by sampling points in a grid-like pattern, which serve as prompts for SAM2. For every segmented entity, we compute RADIO features, repeating this process for all frames. When a new segmentation is detected, its RADIO features are compared to those from the previous frame; if the distance is below a threshold, the entities are labeled as the same object moving between frames, and the frame number and initial location are recorded.
This information is then used to run the SAM2 video tracking pipeline for each object, using its initial position as the prompt. Finally, segmentation maps are postprocessed by merging overlapping entities and replacing frequently overridden ones with the most representative entity. Despite the overhead of running segmentation for each entity and subsequent postprocessing, our method significantly outperforms the traditional SAM2 approach of prompting all objects simultaneously.

\vspace{-5pt}
\subsection{\methodName{}}\label{sec:hierasurg}
We define HieraSurg as a pair of diffusion models based on the CogVideoX-2B \cite{yang2025cogvideox} architecture-a latent Diffusion Transformer (DiT) \cite{peebles_scalable_2023}-which accepts textual prompts to generate videos at various resolutions and frame rates. HieraSurg consists of: HieraSurg-\S2M{} (Semantic to Map), tasked with generating a plausible evolution of the surgical scene in the domain of segmentation maps, and HieraSurg-\M2V{} (Map to Video), which completes the task by bringing said segmentation maps to the video-space. 

\vspace{5pt} \noindent \textbf{HieraSurg-\S2M{}:} this model generates a sequence of $F$ segmentation maps. It takes as input the first video frame $y_1$, its corresponding segmentation map $y_1^{seg}$, and the sequence of the next $F$ phases $ph_i$ and action triplets $tr_i$. Instead of encoding the video via the standard encoder $\mathcal{E}$, we perform a temporally dense latent encoding to preserve individual frame details. We do so by considering a video as a batch of $F$ 1-frame videos and encoding them separately, thus obtaining a latent $z \in \mathbb{R}^{F\times H'\times W'\times d} $, where the frame dimension is $F$ instead of $F'$, $F'$ being the temporal dimension of latents when encoding a video of length $F$ using $\mathcal{E}$.
 Although each segmentation map is naturally a single-channel 2D representation, we convert them to color space before feeding them to the VAE so that their distribution better matches CogVideoX's pretraining data. We inject conditioning along two axes by encoding the initial segmentation frame and the first video frame through the VAE, as $z_y^{seg} = \mathcal{E}(y_1^{seg}), z_y =  \mathcal{E}(y_1)\in \mathbb{R}^{1\times H'\times W'\times d}$, respectively, and then repeating these over the frame dimension to match $z$. They are finally concatenated and provided as input to the denoiser model $\epsilon_\theta$ as: $$z_{in} = [z,z_y^{seg},z_y] \in \mathbb{R}^{F\times H'\times W'\times 3d} $$ 
To condition on $ph_i$ and $tr_i$, we encode each separately and then condense them along the time dimension using a 1D convolution followed by average pooling. We experimented with a learnable embedding layer (Label Embedding) and a pre-trained model like PeskaVLP \cite{yuan_procedure-aware_2024}, where phases and triplets are first mapped to textual representations before embedding. The resulting encoded phase and triplet information are then concatenated with the timestep $t$ encoding and injected into every transformer block.
Diffusion models operate in a continuous space, which means that \S2M{} produces segmentation maps with slightly varied values-an undesirable outcome for accurate segmentation. To address this, we apply K-Means clustering to the set of output colours, finding the optimal number of clusters using an algorithmic elbow method. We then map all colours within each cluster to their centre, resulting in a discrete representation which can be easily mapped to single-channel integer space.

\vspace{5pt} \noindent \textbf{HieraSurg-\M2V{}:} It generates a video sequence of varying duration given an input segmentation map $c$ and the first frame of the video $y_1$. We allow for either the segmentation map to have the same number of frames as the output video or to have a lower amount in the case of high frame-rate video generation. To encode the segmentation maps, we process them through an encoder $\mathcal{E}_{seg}$, made up of a series of 3D residual blocks, each one downsampling the spatial resolution while keeping the temporal dimension intact.\\  
We further add temporally-aware sinusoidal positional embeddings to $\mathcal{E}_{seg}(c)$, which is still a spatially compressed 3D space, before flattening the representation into a 1D sequence $H^{seg}_0$ of length $T_{seg}$. $H^{seg}_0$ is merged into the main flow of the network by way of attention in each of the transformer blocks, where it is further processed:
\begin{align*}
H_{cat} &= [H_{i} ; H^{seg}_i]  \quad\quad Q = W_q H_{cat} \quad K = W_k H_{cat} \quad V = W_v H_{cat} \\
Z &= \text{softmax}\left(\frac{Q K^T}{\sqrt{d_k}}\right)V \quad\quad\quad [H_{i+1}, H^{seg}_{i+1}] = \text{split}(W_o Z, T) 
\end{align*}
Where $H_{cat}$ is the concatenation of the hidden states $H_i$ passing through the block $i$ together with $H^{seg}_i$, $d_k$ is the key vectors dimension, and $W_q, W_k, W_v, W_o$ are learnable weight matrices.
Similarly to \S2M{}, we encode the initial video frame by passing it through the VAE as a single image before stacking it together with the noised latents.

%% file: chapters/3_results.tex
\section{Experiments}
\begin{figure}[tb]
\centering
\includegraphics[width=0.9\textwidth]{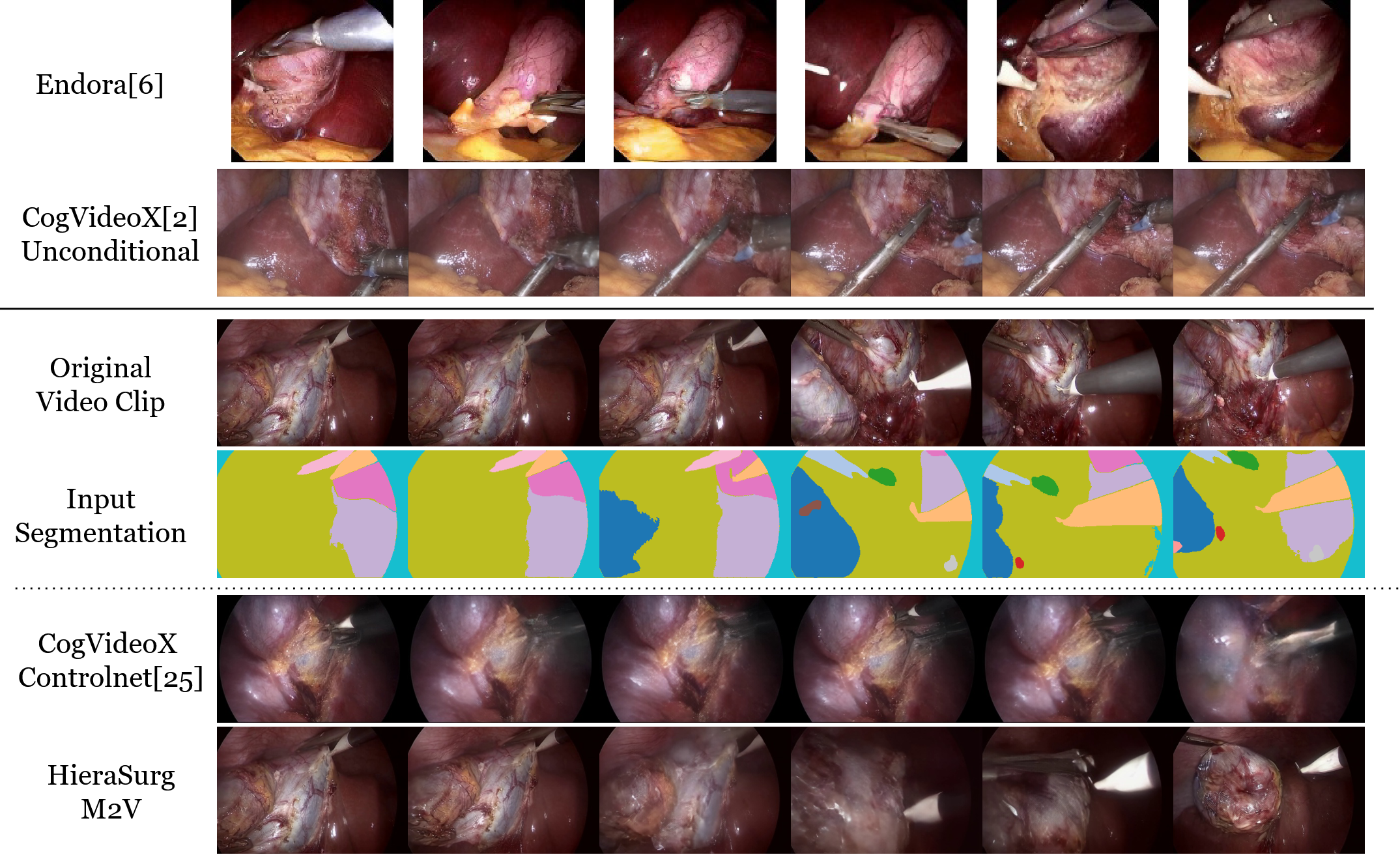}
\caption{Visual comparison of the first 6 seconds generated by different models. For the conditional models, the input segmentation maps are given.}
\label{fig:map2vid}
\end{figure}
\noindent \textbf{Datasets and preprocessing}: We employ Cholec80 \cite{twinanda_endonet_2017}, a collection of 80 cholecystectomy surgeries performed by 13 surgeons acquired as high-resolution videos at 25 FPS with phase annotations; and CholecT45 \cite{nwoye_rendezvous_2022}, a subset containing 45 videos annotated with 100 triplet classes at 1 FPS. 65 videos (41 from CholecT45) are extracted, and the automatic segmentation pipeline is run on overlapping 16-second long clips sampled at 1 FPS. Two videos are held out as a test set. We  cropped and resized the original videos from 854x480 to 384x256.

\vspace{5pt} \noindent \textbf{Evaluation Setting} We evaluate the \textit{visual quality} of the generated videos using Frechet Video Distance (FVD) \cite{unterthiner_fvd_2019}, Frechet Inception Distance (FID) \cite{heusel2017gans} and FID using PeskaVLP \cite{yuan_procedure-aware_2024} as a feature extractor. The \textit{reconstruction ability} is evaluated using the Structural Similarity Index Measure (SSIM). Metrics are computed over 1024 generated samples against the same number of real data points. To validate \textit{segmentation adherence}, we introduce a detector agreement metric. After training a YOLOV8 \cite{varghese_yolov8_2024} model to recognize surgical tools, tool detection is run on all frames of two videos. By matching bounding box predictions, we compute the ratio of objects in the real video that are also found in the generated one(Hit Rate Real), the ratio of generated objects that are found in the real videos (Hit Rate Gen), as well as the mean IoU of matching predictions.

\renewcommand\cellset{\renewcommand\arraystretch{0.5}%
\setlength\extrarowheight{0pt}}
\renewcommand{\arraystretch}{1.1}
\begin{table}[tb]
    \centering
    \caption{Quantitive Comparison of HieraSurg compared to unconditional and conditional baselines. \S2M{}+M2V{} denotes the full \methodName{} pipeline. \textsuperscript{*} Values reported on the paper. \textsuperscript{+} Cholec80 model evaluated on our test split.}
\begin{tabular}{p{2.5cm} p{2cm} >{\centering\arraybackslash}p{0.8cm} >{\centering\arraybackslash}p{0.8cm} >{\centering\arraybackslash}p{1.6cm} >{\centering\arraybackslash}p{0.8cm} >{\centering\arraybackslash}p{0.8cm} >{\centering\arraybackslash}p{0.9cm} >{\centering\arraybackslash}p{0.7cm}}
            \toprule
         & & \multicolumn{3}{c}{Fidelity Metrics $(\downarrow)$} & \multicolumn{4}{c}{Reconstruction Metrics $(\uparrow)$} \\
        \cmidrule(lr){3-5} \cmidrule(lr){6-9}
        Model & Conditioning & FVD & FID & \makecell{PeskaVLP \\ FID}  & \makecell{HR \\ Real} & \makecell{HR \\ Gen}& MIoU & SSIM \\
        \midrule
        \multicolumn{9}{c}{FPS Rate: 1} \\
        \arrayrulecolor{black!30}\midrule
          Endora \cite{li_endora_2024}\textsuperscript{+} & - & 815.8 & 105.2 & 60.3 & - & - & - & 0.16  \\
          CogVideoX \cite{yang2025cogvideox} & - & 443.1 & 79.6 & 35.5 & 0.007 & 0.016 & 0.596 & 0.28 \\
        \arrayrulecolor{black!30}\midrule
        VISAGE \cite{yeganeh_visage_2025}\textsuperscript{*} & Triplet & 1780 & - & - & - & - & - & 0.56  \\     
         \makecell[l]{CogVideoX \\ControlNet \cite{zhang_adding_2023}} & \makecell[l]{GT Seg. \\ Edges} & 640.9 & 82.6 & 28.7 & 0.028 & 0.061 & 0.616 & 0.38 \\
        \M2V{} RADIO & GT Seg. & 640.8 & 72.7 & 36.3 & 0.207 & 0.406 & 0.682 & 0.40  \\
        \M2V{} VAE & GT Seg. & 351.4 & 48.9 & 22.1 & \textbf{0.321} & \textbf{0.476} & \textbf{0.745} & \textbf{0.49}  \\
        \S2M{}+\M2V{} VAE & Pred Seg. & \textbf{312.4} & \textbf{47.1} & \textbf{17.2} & 0.137 & 0.270 & 0.697 & 0.44  \\
        \arrayrulecolor{black}\midrule   
        \multicolumn{9}{c}{FPS Rate: 8} \\
        \arrayrulecolor{black!30}\midrule
        \M2V{} RADIO & GT Seg. & 1202.4 & 71.7 & 33.5 & 0.076 & 0.157 & 0.635 & 0.35  \\
        \M2V{} VAE & GT Seg. & \textbf{276.2} & \textbf{22.4} & \textbf{12.9} & \textbf{0.358} & \textbf{0.447} & 0.806 & \textbf{0.53}  \\
        \S2M{}+\M2V{} VAE & Pred Seg. & 278.0 & 24.1 & 15.1 & 0.218 & 0.311 & \textbf{0.867} & \textbf{0.53}  \\
        \arrayrulecolor{black}\bottomrule      
    \end{tabular}
    \label{tab:results}
\end{table}
\renewcommand{\arraystretch}{1}

\vspace{5pt} \noindent \textbf{Results}
We first validate the ability of HieraSurg-\M2V{} to synthesize 1FPS 16-second videos. After finetuning a CogVideoX-2B text-to-video model to generate unconditional surgical videos by training on zeroed-out captions, the resulting weights are used as a starting point for all \M2V{} variants.
We experiment with providing \M2V{} the initial image through different encoding methods, either RADIO or VAE-based. We further train and evaluate the best-performing model on 8FPS 6-second videos.
The condition-following ability of \M2V{} is compared to ControlNet \cite{zhang_adding_2023}, a popular architecture to inject control into an LDM. In particular, we take the implementation of ControlNet for CogVideoX, which uses Canny edges, and finetune it on edges extracted from our segmentation maps.
The results (\textit{cf.} \autoref{tab:results}) show improved values across fidelity metrics by \M2V{}, especially in the 8FPS setting. Our method's ability to follow guidance is also clearly superior compared to the ControlNet baseline, the best results being obtained when the initial frame is provided through the VAE encoding. Visual results are found in \autoref{fig:map2vid}.
\begin{figure}[tb]
\centering
\includegraphics[width=0.98\textwidth]{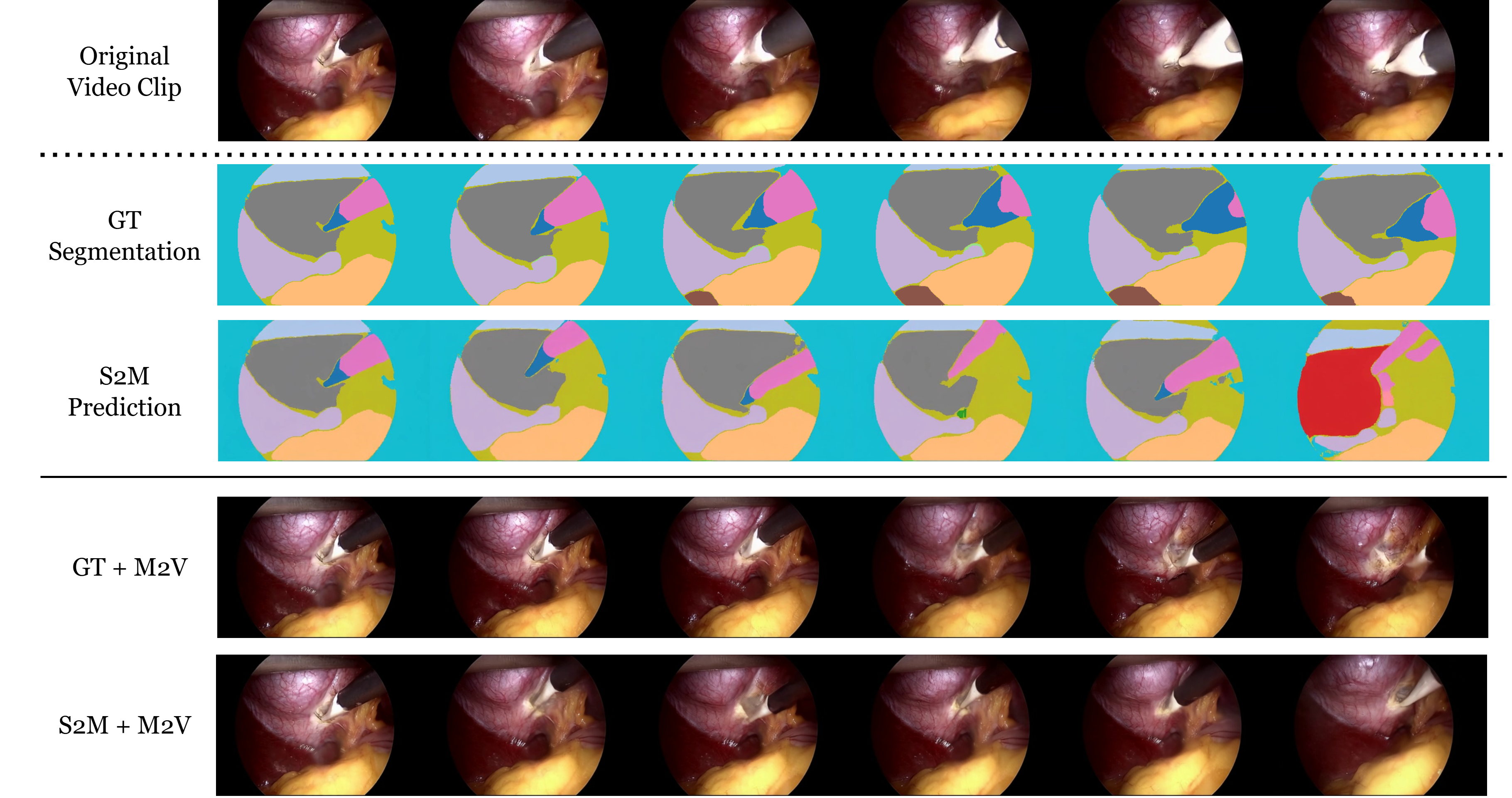}
\caption{Example output of the full HieraSurg pipeline, comparing the usage of \M2V{} with the ground truth segmentation to when providing the \S2M{} output to it}
\label{fig:hierasurg_full_res}
\end{figure}
Next, we train HieraSurg-\S2M{} at 1FPS, allowing the prediction of segmentation maps for the next 16 seconds. We experimented with encoding the phase and triplet information using both pretrained and learned embeddings. In \autoref{tab:ablation}, we show that the former method gives better results. Thus it is used in further experiments.
The coherency of the generated predictions is validated by providing the output of \S2M{} to \M2V{} in both the 1FPS and the 8FPS setting.
When using predicted segmentation maps, we notice that visual quality is only slightly affected while the ability to follow the real-world evolution of the surgery is diminished, as expected. Detector agreement is significantly lower in the 1FPS setting due to the inherent complexity in predicting 16 seconds into the future compared to 6. Complete quantitative evaluations are found in \autoref{tab:results}, while a visual comparison of the full pipeline is found in \autoref{fig:hierasurg_full_res}.
\begin{table}[tb]
    \centering
    \caption{Ablation Study of \S2M{},  with the full HieraSurg pipeline @1FPS}
    \begin{tabular}{c|cc|cccc}
        \toprule
        \makecell[c]{ Temporally\\[-1pt] Dense Latent} & \makecell[c]{ Phase/Triplet\\[-1pt] Cond.} & \makecell[c]{ Cond \\[-1pt] Method} & FVD $(\downarrow)$ & FID$(\downarrow)$ & \makecell[c]{HR\\Real $(\uparrow)$} & SSIM $(\uparrow)$ \\
        \midrule
        \checkmark & \ding{55} & - & 378.6 & 52.5 & 0.113 & \textbf{0.44} \\
          \ding{55} & \checkmark &  PeskaVLP & 470.3 & 50.4 & 0.104 & 0.38 \\
        \checkmark & \checkmark & Label Emb. & 389.6 & 53.2 & 0.119 & 0.43 \\
        \checkmark & \checkmark & PeskaVLP &  \textbf{312.4} & \textbf{47.1} & \textbf{0.137} & \textbf{0.44}\\
        \bottomrule
        
    \end{tabular}
    \label{tab:ablation}
\end{table}
An ablation study of S2M{} is provided in \autoref{tab:ablation}, showing that the presence of surgical information does help and that supplying robust textual encodings of phase and triplet, as given by PeskaVLP, is beneficial over letting the model learn them from scratch. Furthermore, we validate the latent encoding choice by showing that a temporally compressed latent space yields segmentation maps that are harder to parse for the second stage model. Most notably, the ablated first-stage outputs show blurrier edges and a higher amount of colour variability in areas that are supposed to be uniform, making the K-Means procedure less effective.

%% file: chapters/4_conclusion.tex
\section{Conclusion}
This work proposes \methodName{} for high-quality and controllable surgical video generation. HieraSurg-\M2V{} is able to synthesize videos with impressive visual quality, which strictly adheres to the conditioning given through segmentation maps. Thanks to HieraSurg-\S2M{}, we are further able to generate completely novel videos starting from only an initial frame and surgical phase/triplet information, offering a glimpse into one of the possible developments of the scene.
This paves the way for its usage in multiple approaches, either only as \M2V{} to visualize the development of a known scene or as the full HieraSurg for a free-form generation task.
Nevertheless, the predictive capacity of our pipeline is heavily reliant on the quality of the segmentation maps and the ability of the first-stage component to keep track of and contribute a valid trajectory for the entities. Further developments would be focused on improving the first-stage model, ideally leveraging an even higher amount of semantic information, which could constrain the fan of conceivable trajectories, enabling a more reliable anticipation of the future state of the scene.